%% file: main.tex
\documentclass[runningheads]{llncs}
\usepackage{graphicx}
\usepackage{comment}
\usepackage{amsmath,amssymb} 
\usepackage{dsfont} 
\usepackage{color}
\usepackage{pifont}
\usepackage[dvipsnames]{xcolor}
\usepackage{array}
\usepackage{multirow}
\usepackage{mathtools}
\usepackage[]{algorithm2e}

\newcommand{\eg}{\emph{e.g.~}}
\newcommand{\ie}{\emph{i.e.~}}

\newcommand{\myparagraph}[1]{\noindent\textbf{#1}}

\newcommand{\methodName}{{CuMix} }
\newcommand{\methodNameFull}{\textbf{Cu}rriculum \textbf{Mix}up for recognizing unseen categories in unseen domains}

\begin{document}
\pagestyle{headings}
\mainmatter
\def\ECCVSubNumber{4355}  

\title{Towards Recognizing Unseen Categories\\ in Unseen Domains}


\titlerunning{Towards Recognizing Unseen Categories in Unseen Domains}
%
\author{Massimiliano Mancini\inst{1,2}\orcidID{0000-0001-8595-9955} \and
\\Zeynep Akata\inst{2}\orcidID{0000-0002-1432-7747} \and
Elisa Ricci\inst{3,4}\orcidID{0000-0002-0228-1147} \and
Barbara Caputo\inst{5,6}\orcidID{0000-0001-7169-0158}}
\authorrunning{M. Mancini et al.}
%
\institute{
Sapienza University of Rome, \and University of T\"ubingen, \\\and University of Trento, \and Fondazione Bruno Kessler, \\ \and Politecnico di Torino, \and Italian Instituite of Technology
\\ \email{mancini@diag.uniroma1.it}}
\maketitle

\begin{abstract} 
Current deep visual recognition systems suffer from severe performance degradation when they encounter new images from classes and scenarios unseen during training. Hence, the core challenge of Zero-Shot Learning (ZSL) is to cope with the \textit{semantic-shift} whereas the main challenge of Domain Adaptation and Domain Generalization (DG) is the \textit{domain-shift}. 
While historically ZSL and DG tasks are tackled in isolation, this work develops with the ambitious goal of solving them jointly, i.e. by recognizing \emph{unseen visual concepts in unseen domains}. We present \methodName (\methodNameFull), a holistic algorithm to tackle ZSL, DG and ZSL+DG. The key idea of \methodName is to simulate the test-time domain and semantic shift using images and features from unseen domains and categories generated by \textit{mixing up} the multiple source domains and categories available during training. Moreover, a curriculum-based mixing policy is devised to generate increasingly complex training samples.
Results on standard ZSL and DG datasets and on ZSL+DG using the DomainNet benchmark demonstrate the effectiveness of our approach.

\end{abstract}

\section{Introduction}
\input{intro.tex}

\section{Related Works}
\input{related.tex}

\section{Method}
\input{method.tex}

\section{Experimental results}
\input{experiments.tex}

\section{Conclusions}
\input{conclusions.tex}



%
%
\bibliographystyle{splncs04}
\bibliography{egbib}

\section*{\Large Appendix}
\appendix

\section{Hyperparameter choices}
\input{suppmat/hyperparams}

\section{ZSL+DG: analysis of additional baselines}
\input{suppmat/zsl-dg}

\section{ZSL+DG: ablation study}
 In order to further investigate our design choices on the ZSL+DG setting, we conducted experiments on a challenging scenario where we consider just two domains as sources, i.e. Real and Painting. The results are shown in Table \ref{tab:ablation-zsldg}. On average our model improves SPNet by 2\% and SPNet + Epi-FCR by 1.1\%. Our approach without curriculum largely outperforms standard image-level mixup \cite{zhang2017mixup} (more than 2\%). Applying mixup at both feature and image level but without curriculum is effective but achieves still lower results with respect to our CuMix strategy (as in Tab. 2).  Interestingly, if we apply the curriculum strategy but switching the order of semantic and domain mixing (CuMix reverse), this achieves lower performances with respect to CuMix, which considers domain mixing harder than semantic ones. This shows that, in this setting, it is important to correctly tackle intra-domain semantic mixing before including inter-domain ones.
 
 \begin{table*}[t]
           \label{tab:ablation-zsldg}
 			\caption{Results on DomainNet dataset with \textit{Real-Painting} as sources and ResNet-50 as backbone.} 
 			 		\centering
 		\begin{tabular}{ l | c  c  c  c | c}
 		Method/Target & Clipart& Infograph& Sketch& Quickdraw& Avg.\\\hline 
 SPNet& 21.5$\pm{0.6}$& 14.1$\pm{0.2}$& 17.3$\pm{0.3}$& 4.8$\pm{0.4}$& 14.4\\
 Epi-FCR+SPNet& 22.5$\pm{0.5}$& 14.9$\pm{0.7}$& 18.7$\pm{0.6}$& \textbf{5.6}$\pm{0.4}$& 15.4\\
 \hline
 MixUp img only & 21.2$\pm{0.4}$& 14.0$\pm{0.7}$& 17.3$\pm{0.3}$& 4.8$\pm{0.1}$& 14.3\\
 MixUp two-level& 22.7$\pm{0.3}$& 16.5$\pm{0.4}$& 19.1$\pm{0.4}$& 4.9$\pm{0.3}$& 15.8\\
 CuMix reverse& 22.9$\pm{0.3}$& 15.8$\pm{0.2}$& 18.2$\pm{0.3}$& 4.8$\pm{0.5}$& 15.4\\
 \hline
 CuMix & \textbf{23.7$\pm{0.3}$}& \textbf{17.1$\pm{0.2}$}& \textbf{19.7$\pm 0.3$}& {5.5$\pm{0.3}$}& \textbf{16.5} \\
 		\end{tabular}
 		\label{tab:ablation-study-pacs}
 \end{table*}

\section{ZSL results}
\input{suppmat/zsl-ablation}
\end{document}

%% file: intro.tex
Despite their astonishing success in several applications \cite{girshick2015fast,redmon2016you}, deep visual models perform poorly for the classes and scenarios that are unseen during training. 
Most existing approaches are based on the assumptions that (a) training and test data come from the same underlying distribution, i.e. domain shift, and (b) the set of classes seen during training constitute the only classes that will be seen at test time, i.e. semantic shift.  
These assumptions rarely hold in practice and, in addition to depicting different semantic categories, training and test images may differ significantly in terms of visual appearance in the real world.

To address these limitations, research efforts have been devoted to designing deep architectures able to cope with varying visual appearance \cite{csurka2017domain} and with novel semantic concepts \cite{xian2018zeroshotgood}.
In particular, the domain-shift problem \cite{khosla2012undoing} has been addressed by proposing domain adaptation (DA) models \cite{csurka2017domain} that assume the availability target domain data during 
training. To circumvent this assumption, a recent trend has been to move to more complex scenarios where the adaptation problem must be either tackled online \cite{hoffman2014continuous,mancini2018kitting}, with the help of target domain descriptions \cite{mancini2019adagraph}, auxiliary data \cite{peng2018zero} 
or multiple source domains \cite{mancini2019inferring,mancini2018boosting,roy2019unsupervised}. For instance, domain generalization (DG) methods \cite{li2017deeper,li2019episodic,carlucci2019domain} aim to learn domain-agnostic prediction models and to generalize to any unseen target domain.

Regarding semantic knowledge, 
multiple works have 
designed approaches for extending deep architectures to handle new categories and new tasks. For instance, continual learning methods \cite{DeLange-CL-survey} attempt to sequentially learn new tasks 
while retaining previous knowledge, tackling the catastrophic forgetting issue. Similarly, in open-world recognition \cite{bendale2015towards} the goal is to detect 
unseen categories and successfully incorporate them into the model. Another research thread is Zero-Shot Learning (ZSL) \cite{xian2018zeroshotgood}, where the goal is to recognize objects unseen during training given external information about the novel classes provided in forms of semantic attributes \cite{lampert2013awa}, visual descriptions \cite{akata2015evaluation} or word embeddings \cite{mikolov2013efficient}.

\begin{figure}[tb]
  \centering
  \includegraphics[width=0.94\columnwidth,trim=0 5pt 0 0,clip]{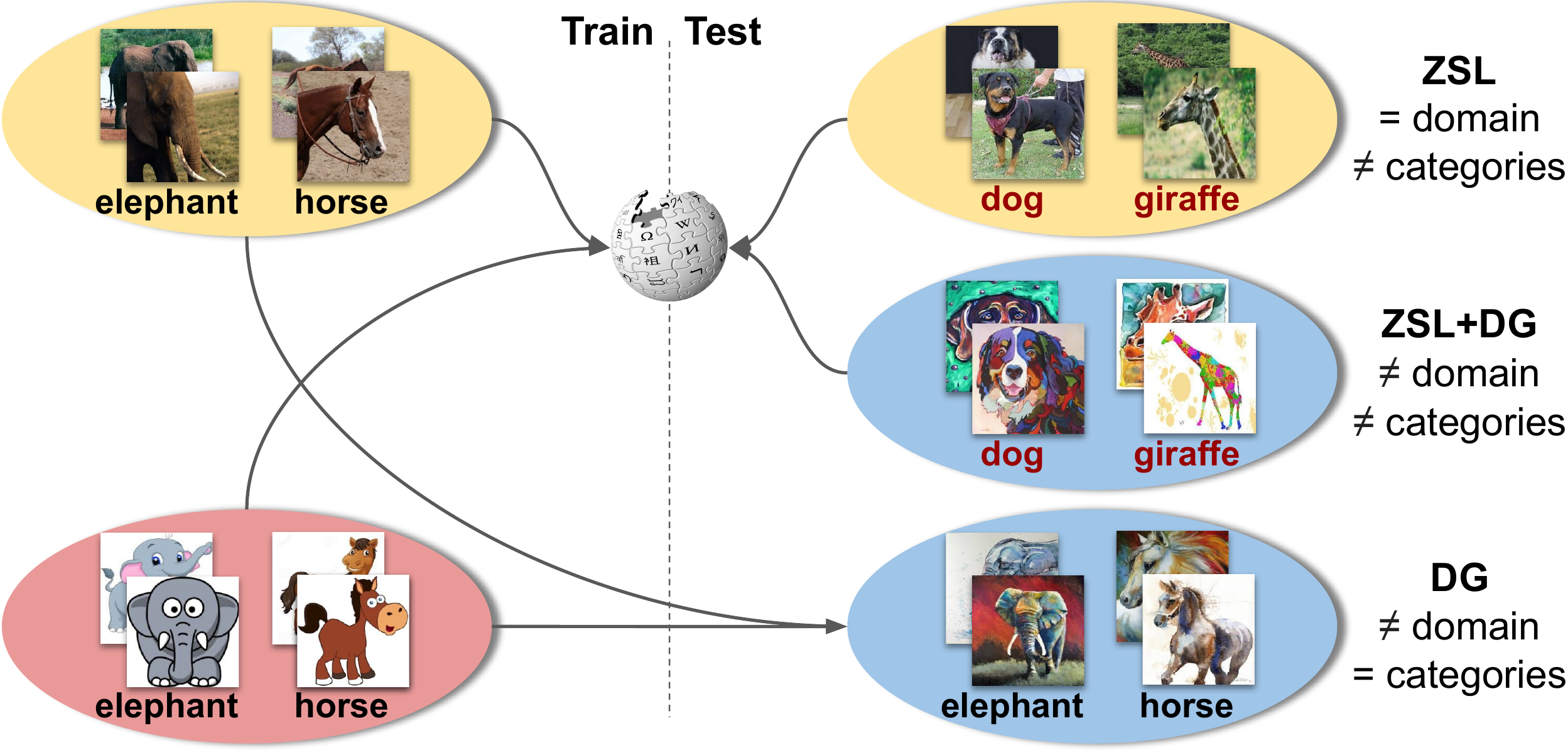}
   \caption{Our ZSL+DG problem. During training we have images of multiple categories (\eg \textit{elephant},\textit{horse}) and domains (\eg \textit{photo}, \textit{cartoon}). At test time, we want to recognize unseen categories (\eg \textit{dog}, \textit{giraffe}), as in ZSL, in unseen domains (\eg \textit{paintings}), as in DG, exploiting side information describing seen and unseen categories. }
  \label{fig:teaser}
 \end{figure}

Despite these significant efforts, an open research question is whether we can tackle the two problems jointly. {Indeed, due to the large variability of visual concepts in the real world, in terms of both semantics and acquisition conditions, it is impossible to construct a training set capturing such variability. This calls for a holistic approach addressing them together.} Consider for instance the case depicted in Fig.~\ref{fig:teaser}. A system trained to recognize elephants and horses from realistic images and cartoons might be able to recognize the same categories in another visual domain, like art paintings (Fig.~\ref{fig:teaser}, bottom) or it might be able to describe other quadrupeds in the same training visual domains (Fig.~\ref{fig:teaser}, top). On the other hand, how to deal with the case where new animals are shown in a new visual domain is not clear.

To our knowledge, our work is the first attempt to answer this question, proposing a method that is able to \textit{recognize unseen semantic categories in unseen domains}. In particular, our goal is to jointly tackle ZSL and DG (see Fig.\ref{fig:teaser}). ZSL algorithms usually receive as input a set of images with their associated semantic descriptions, and learn the relationship between an image and its semantic attributes. Likewise, DG approaches are trained on multiple source domains and at test time are asked to classify images, assigning labels within the same set of source categories but in an unseen target domain. 
Here we want to address the scenario where, during training, 
\emph{we are given a set of images of multiple domains and semantic categories and our goal is to build a model able to recognize images of unseen concepts, as in ZSL, in unseen domains, as in DG.}

To achieve this, we need to address challenges usually not present  
when these two classical tasks, i.e. ZSL and DG, are considered in isolation. For instance, while in DG we can rely on the fact that the multiple source domains permit to disentangle semantic and domain-specific information, in ZSL+DG we have no guarantee that the disentanglement will hold for the unseen semantic categories at test time. Moreover, while in ZSL it is reasonable to assume that the learned mapping between images and semantic attributes will generalize also to test images of the unseen concepts, 
in ZSL+DG we have no guarantee that this will happen for images of unseen domains. 

{To overcome these issues, during training we 
simulate
both the semantic and the domain shift we will encounter at test time. Since explicitly generating images of unseen domains and concepts is an ill-posed problem, we sidestep this issue and we synthesize unseen domains and concepts by interpolating existing ones. To do so, we revisit the \textit{mixup} \cite{zhang2017mixup} algorithm as a tool to obtain partially unseen categories and domains. Indeed, by randomly mixing samples of different categories we obtain new samples which do not belong to a single one of the available categories during training. Similarly, by mixing samples of different domains, we obtain new samples which do not belong to a single source domain available during training.

Under this perspective, mixing samples of both different domains and classes allows to obtain samples that cannot be categorized in a single class and domain of the one available during training, thus they are \textit{novel} both for the semantic and their visual representation. Since higher levels of abstraction contain more task-related information, we perform \textit{mixup} at both image and feature level, showing experimentally the need for this choice. {Moreover, we introduce  a  curriculum-based  mixing  strategy to generate increasingly complex training samples. } 
We show that our \methodName (\methodNameFull) model obtains state-of-the-art performances in both ZSL and DG in standard benchmarks and it can be effectively applied to the combination of the two tasks, recognizing unseen categories in unseen domains.\footnote{The code is available at \textit{\url{https://github.com/mancinimassimiliano/CuMix}}}} 

To summarize, our contributions are as follows.
(i) We introduce the ZSL+DG scenario, a first step towards recognizing unseen categories in unseen domains.
(ii) Being the first holistic method able to address ZSL, DG, and the two tasks together, our method is based on simulating new domains and categories during training by mixing the available training domains and classes both at image and feature level. The mixing strategy becomes increasingly more challenging during training, in a curriculum fashion.
(iii) Through our extensive evaluations and analysis, we show the effectiveness of our approach in all three settings: namely ZSL, DG and ZSL+DG.

%% file: related.tex
\myparagraph{Domain Generalization (DG).} Over the past years the research community has put considerable efforts into developing methods to contrast the domain shift. Opposite to domain adaptation \cite{csurka2017domain}, where it is assumed that target data are available in the training phase, the key idea behind DG is to learn a domain agnostic model to be applied to any unseen target domain. 

Previous DG methods can be broadly grouped into four main categories.
The first category comprises methods which attempt to learn
domain-invariant feature representations \cite{muandet2013domain} by considering specific alignment losses, such as maximum mean discrepancy (MMD), adversarial loss \cite{li2018domainadv} or self-supervised losses \cite{carlucci2019domain}. 
The second category of methods \cite{li2017deeper,khosla2012undoing} 
develop from the idea of creating deep architectures where both domain-agnostic and domain-specific parameters are learned on source domains. After training, only the domain-agnostic part is retained and used for processing target data. 
The third category devises specific optimization strategies or training procedures in order to enhance the generalization ability of the source model to unseen target data. For instance, in \cite{li2018learning} a meta-learning approach is proposed for DG. Differently, in \cite{li2019episodic} an episodic training procedure is presented to learn models robust to the domain shift. 
The latter category comprises methods which introduce 
data and feature augmentation strategies to synthesize novel samples and improve the generalization capability of the learned model \cite{shankar2018generalizing,volpi2018generalizing,volpi2019addressing}. These strategies are mostly based on adversarial training \cite{shankar2018generalizing,volpi2018generalizing}.

Our work is related to the latter category since we also generate synthetic samples with the purpose of learning more robust target models. However, differently from previous methods, we specifically employ mixup to perturb feature representations.
Recently, works have considered mixup in the context of domain adaptation {\cite{xu2019adversarial}} to \eg reinforce the judgments of a domain discrimination. However, we employ mixup from a different perspective \ie simulating semantic and domain shift we will encounter at test time. To this extent, we are not aware of previous methods using mixup for DG and ZSL.

\myparagraph{Zero-Shot Learning (ZSL).} 
Traditional ZSL approaches attempt to learn a projection function mapping images/visual features to a semantic embedding space where  classification is performed. This idea is achieved by directly predicting image attributes e.g. \cite{lampert2013awa} or by learning a linear mapping through margin-based objective functions \cite{akata2013label,akata2015evaluation}. Other approaches explored the use of non-linear multi-modal embeddings \cite{xian2016latent}, intermediate projection spaces \cite{zhang2015zero,zhang2016zero} or similarity-based interpolation of base classifiers \cite{changpinyo2016synthesized}. Recently, various methods tackled ZSL from a generative point of view considering Generative Adversarial Networks  \cite{xian2018feature}, Variational Autoencoders (VAE) \cite{schonfeld2019generalized} or both of them \cite{xian2019fvaegan}. While none of these approaches explicitly tackled the domain shift, i.e. visual appearance changes among different domains/datasets, various methods proposed to use domain adaptation technique, e.g. to refine the semantic embedding space, aligning semantic and projected visual features \cite{schonfeld2019generalized} or, in transductive scenarios, to cope with the inherent domain shift existing among the appearance of attributes in different categories \cite{Kodirov_2015_ICCV,fu2015transductive,gan2016learning}. For instance, in \cite{schonfeld2019generalized} a distance among visual and semantic embedding projected in the VAE latent space is minimized. In \cite{Kodirov_2015_ICCV} the problem is addressed through a regularised sparse coding framework, while in \cite{fu2015transductive} a multi-view hypergraph label propagation framework is introduced.

Recently, works have considered also coupling ZSL and DA in a transductive setting. For instance, in \cite{zhuo2019unsupervised} a semantic guided discrepancy measure is employed to cope with the asymmetric label space among source and target domains. In the context of image retrieval, multiple works addressed the sketch-based image retrieval problem \cite{yelamarthi2018sketch,Dutta_2019_CVPR}, even across multiple domains. In \cite{thong2019open} the authors proposed a method to perform cross-domain image retrieval by training domain-specific experts. While these approaches integrated DA and ZSL, none of them considered the more complex scenario of DG, where no target data are available.

%% file: method.tex
In this section, we first formalize the Zero-Shot Learning under Domain Generalization (ZSL+DG). We then describe our approach, \methodName, which, by performing curriculum learning through mixup, simulates the domain- and semantic-shift the network will encounter at test time, and can be holistically applied to ZSL, DG and ZSL+DG. 

\subsection{Problem Formulation}
In the ZSL+DG problem, the goal is to recognize unseen categories (as in ZSL) in unseen domains (as in DG). Formally, let $\mathcal{X}$ denote the input space (e.g. the image space), $\mathcal{Y}$ the set of possible classes and $\mathcal{D}$ the set of possible domains. During training, we are given a set  $\mathcal{S}=\{(x_i,y_i,d_i)\}_{i=1}^n$ where $x_i\in\mathcal{X}$, $y_i \in \mathcal{Y}^s$ and $d_i \in \mathcal{D}^s$. Note that $\mathcal{Y}^s\subset \mathcal{Y}$ and $\mathcal{D}^s\subset \mathcal{D}$ and, as in standard DG, {we have multiple source domains (\ie $\mathcal{D}^s=\cup_{j=1}^m d_j$, with $m>1$) with} different distributions \ie $p_\mathcal{X}(x|d_i)\neq p_\mathcal{X}(x|d_j)$, $\forall i\neq j$.

Given $\mathcal{S}$ our goal is to learn a function $h$ mapping an image $x$ of 
domains $\mathcal{D}^u\subset \mathcal{D}$ to its corresponding label in a set of 
classes $\mathcal{Y}^u\subset \mathcal{Y}$. Note that in standard ZSL, {while the set of train and test domains are shared, \ie $\mathcal{D}^s\equiv\mathcal{D}^u$}, the label sets are disjoint \ie $\mathcal{Y}^s\cap\mathcal{Y}^u \equiv \emptyset$, {thus $\mathcal{Y}^u$ is a set of \textit{unseen} classes. } On the other hand, in DG we have {a shared output space}, \ie $\mathcal{Y}^s\equiv\mathcal{Y}^u$, but a disjoint set of domains between training and test \ie $\mathcal{D}^s\cap\mathcal{D}^u \equiv \emptyset$, 
{thus $\mathcal{D}^u$ is a set of \textit{unseen} domains}. Since the goal of our work is to recognize unseen classes in unseen domains, we unify the settings of DG and ZSL, considering both semantic- and domain-shift at test time \ie $\mathcal{Y}^s\cap\mathcal{Y}^u \equiv \emptyset$ and $\mathcal{D}^s\cap\mathcal{D}^u \equiv \emptyset$.

In the following we divide the function $h$ into three parts: $f$, mapping images into a feature space $\mathcal{Z}$, \ie $f:\mathcal{X}\rightarrow \mathcal{Z}$, $g$ going from $\mathcal{Z}$ to a semantic embedding space $\mathcal{E}$, \ie $g:\mathcal{Z}\rightarrow \mathcal{E}$, and an embedding function {$\omega:\mathcal{Y}^t\rightarrow\mathcal{E}$} 
where $\mathcal{Y}^t\equiv\mathcal{Y}^s$ during training and $\mathcal{Y}^t\equiv\mathcal{Y}^u$ at test time. Note that $\omega$ is a learned classifier for DG while it is a fixed semantic embedding function in ZSL, mapping classes into their vectorized representation extracted from external sources.
Given an image $x$, the final class prediction is obtained as follows:
\begin{equation}
    y^{*} = \text{argmax}_{y} {\omega(y)}^{\intercal}g(f(x)).
\end{equation}
In this formulation, $f$ can be any learnable feature extractor (\eg a deep neural network), while $g$ any ZSL predictor {(\eg a semantic projection layer, as in \cite{xian2019semantic} or a compatibility function among visual features and labels, as in \cite{akata2013label,akata2015evaluation})}. 
The first solution {to address the ZSL+DG problem} could be training a classifier using the aggregation of data from all source domains. In particular, for each sample we could minimize a loss function of the form:
\begin{equation}
    \label{eq:aggregation}
    \mathcal{L}_{\text{AGG}}(x_i,y_i) = \sum_{y\in\mathcal{Y}^s} \ell(\omega(y)^\intercal g(f(x_i)),y_i)
\end{equation}
with $\ell$ an arbitrary loss function, \eg the cross-entropy loss. In the following, we show how we can use the input to Eq.~\eqref{eq:aggregation} to effectively recognize unseen categories in unseen domains.

\subsection{Simulating Unseen Domains and Concepts through \textit{Mixup}}

\begin{figure}[tb]
  \centering
  \includegraphics[width=0.94\columnwidth, trim=0 6pt 0 0,clip]{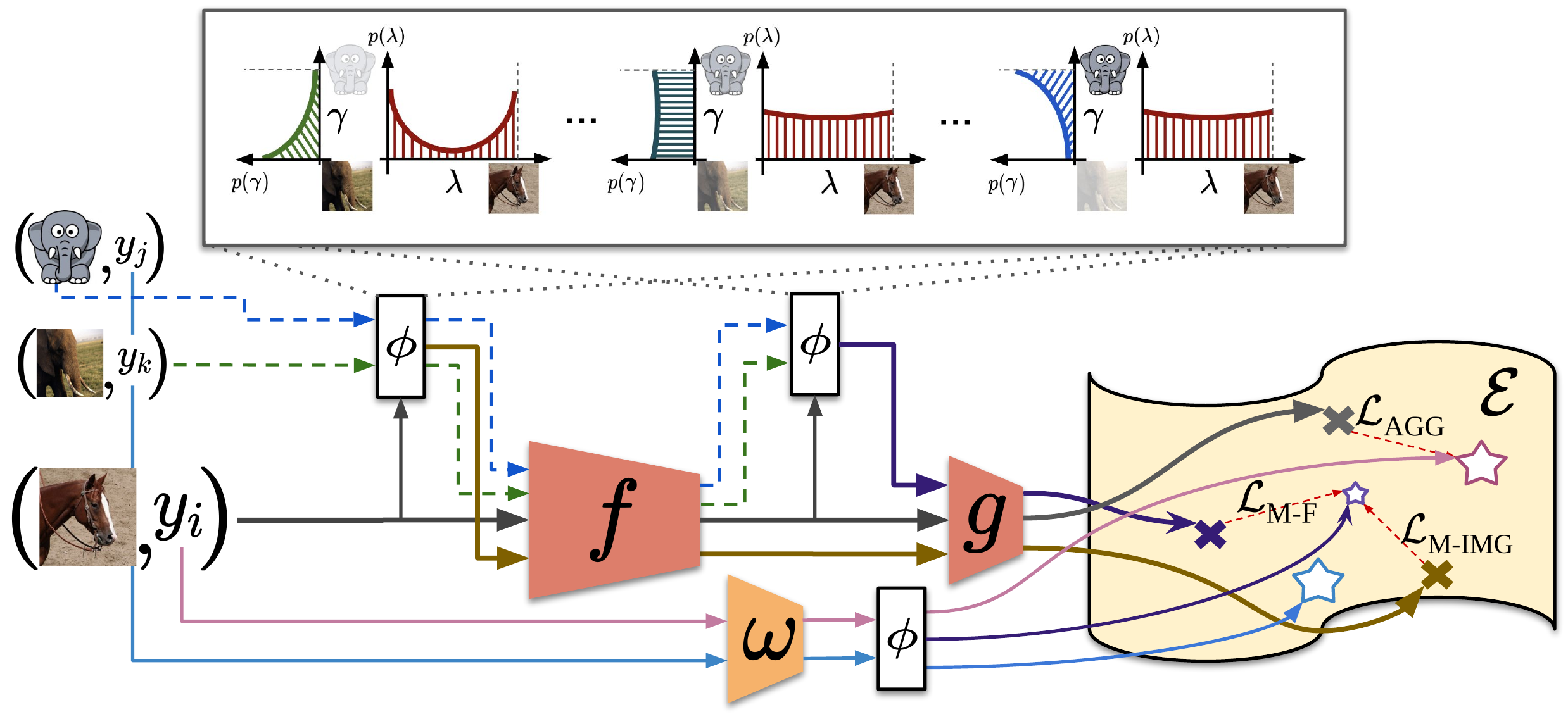} 
  \caption{Our \methodName Framework. Given an image (bottom, \textit{horse}, \textit{photo}), we randomly sample one image from the same (middle, \textit{photo}) and one from another (top, \textit{cartoon}) domain. The samples are mixed through $\phi$ (white blocks) both at image and feature level, with their features and labels projected into the embedding space $\mathcal{E}$ (by $g$ and $\omega$ respectively) and there compared to compute our final objective. Note that $\phi$ varies during training (top part), changing the mixing ratios in and across domains. }
  \label{fig:method}
 \end{figure}
 
The fundamental problem of ZSL+DG is that, during training, we have neither access to visual data associated to categories in $\mathcal{Y}^u$ nor to data of the unseen domains $\mathcal{D}^u$. One way to overcome this issue in ZSL is to generate samples of unseen classes by learning a generative function conditioned on the semantic embeddings in {$\mathcal{W}=\{\omega(y) | y\in\mathcal{Y}^{s}\}$} \cite{xian2018feature,xian2019fvaegan}. However, since no {description} is available for the unseen target domain(s) in $\mathcal{D}^u$, this strategy is not feasible in ZSL+DG. On the other hand, previous works on DG proposed to synthesize images of unseen domains through adversarial strategies of data augmentation  {\cite{volpi2018generalizing,shankar2018generalizing}}. However, these strategies are not applied to ZSL since they cannot easily be extended to generate data for unseen semantic categories 
$\mathcal{Y}^u$. 

To circumvent this issue, we introduce a strategy to simulate, during training, novel domains and semantic concepts by interpolating from the ones available in $\mathcal{D}^s$ and $\mathcal{Y}^s$. Simulating novel domains and classes allows to train the network to cope with both semantic- and domain-shift, the same situation our model will face at test time. Since explicitly generating inputs of novel domains and categories is a complex task, in this work we propose to achieve this goal, by \textit{mixing} images and features of different classes and domains, revisiting the popular \textit{mixup} \cite{zhang2017mixup} strategy. 

In practice, given two elements $a_i$ and $a_j$ of the same space (\eg $a_i,a_j\in\mathcal{X}$), \textit{mixup} \cite{zhang2017mixup} defines a mixing function $\varphi$ as follows: 
\begin{equation}
    \label{eq:phi-mixup}
    \varphi(a_i,a_j) =\lambda\cdot a_i + (1-\lambda) \cdot a_j
\end{equation}
with $\lambda$ sampled from a beta distribution, \ie $\lambda\sim \text{Beta}(\beta,\beta)$, with $\beta$ an hyperparameter. Given two samples $(x_i,y_i)$ and $(x_j,y_j)$ randomly drawn from a training set $\mathcal{T}$, a new loss term is defined as:
\begin{equation}
    \label{eq:mixuploss}
    \mathcal{L}_{\text{MIXUP}}((x_i,y_i),(x_j,y_j)) = \mathcal{L}_{\text{AGG}}(\varphi(x_i,x_j),\varphi(\bar{y}_i,\bar{y}_j))
\end{equation}
where $\bar{y}_i\in \Re^{|\mathcal{Y}^s|}$ is the one-hot vectorized representation of label $y_i$. Note that, when mixing two samples and label vectors with $\varphi$, a single $\lambda$ is drawn and applied within $\varphi$ in both image and label spaces. The loss defined in Eq.\eqref{eq:mixuploss} forces the network to disentangle the various semantic components (\ie $y_i$ and $y_j$) contained in the mixed inputs (\ie $x_i$ and $x_j$) plus the ratio $\lambda$ used to mix them.
This auxiliar task acts as a strong regularizer that helps the network to \eg being more robust against adversarial examples \cite{zhang2017mixup}. Note however that the function $\varphi$ creates input and targets which {do not represent} a single semantic concept in $\mathcal{T}$ but contains characteristics taken from multiple samples and categories, synthesising a \textit{new} semantic concept from the interpolation of existing ones. 

For recognizing unseen concepts in unseen domains at test time, we revisit $\varphi$ to obtain both cross-domain and cross-semantic mixes during training, simulating both semantic- and domain-shift. While simulating the semantic-shift is a by-product of the original \textit{mixup} formulation, here we explicitly revisit $\varphi$ in order to perform cross-domain mixups. In particular, instead of considering a pair of samples from our training set, we consider a triplet $(x_i,y_i,d_i)$, $(x_j,y_j,d_j)$ and $(x_k,y_k,d_k)$. Given $(x_i,y_i,d_i)$, the other two elements of the triplet are randomly sampled from $\mathcal{S}$, with the only constraint that $d_i=d_k,i\neq k$ and $d_j\neq d_i$. In this way, the triplet contains two samples of the same domain (\ie $d_i$) and a third of a different one (\ie $d_j$). Then, our mixing function $\phi$ is defined as follows: 
\begin{equation}
    \label{eq:our-phi}
    \phi(a_i,a_j,a_k) = \lambda a_i + (1-\lambda)( \gamma a_j + (1-\gamma) a_k)
\end{equation}
with $\gamma$ sampled from a Bernoulli distribution $\gamma \sim \mathcal{B}(\alpha)$ and $a$ representing either the input $x$ or the vectorized version of the label $y$, \ie $\bar{y}$. Note that we introduced a term $\gamma$ which allows to perform either intra-domain (with $\gamma=0$) or cross-domain (with $\gamma=1$) mixes.  

To learn a feature extractor $f$ and a semantic projection layer $g$ robust to domain- and semantic-shift, we propose to use  $\phi$ to simulate both samples and features of novel domains and classes during training. Namely, we simulate the semantic- and domain-shift at two levels, i.e. image and class levels. 
 Given a sample $(x_i,y_i,d_i)\in\mathcal{S}$ we define the following loss:
\begin{equation}
    \label{eq:mixuploss-img}
    \mathcal{L}_{\text{M-IMG}}(x_i,y_i,d_i) = \mathcal{L}_{\text{AGG}}(\phi(x_i,x_j,x_k),\phi(\bar{y}_i,\bar{y}_j,\bar{y}_k)).
\end{equation}
where $(x_i,y_i,d_i)$,$(x_j,y_j,d_j)$,$(x_k,y_k,d_k)$ are randomly sampled from $\mathcal{S}$, with $d_i=d_k$ and $d_j\neq d_k$. 
The loss term in Eq. \eqref{eq:mixuploss-img} enforces the feature extractor to effectively process inputs of mixed domains/semantics obtained through $\phi$. Additionally, to also act at classification level, we design another loss which forces the semantic consistency of mixed features in $\mathcal{E}$. This loss term is defined as: 
\begin{equation}
    \label{eq:mixuploss-feats}
    \mathcal{L}_{\text{M-F}}(x_i,y_i,d_i) =  \sum_{y\in\mathcal{Y}^s} \ell\biggl(\omega(y)^\intercal g\bigl(\phi(f(x_i),f(x_j),f(x_k))\bigr),\phi(\bar{y}_i,\bar{y}_j,\bar{y}_k)\biggr)
\end{equation} 
where, as before, $(x_j,y_j,d_j),(x_k,y_k,d_k)\sim \mathcal{S}$, with $d_i=d_k,i\neq k$ and $d_j\neq d_k$ and $\ell$ is a generic loss function \eg the cross-entropy loss. 
This second loss term forces the classifier $\omega$ and the semantic projection layer $g$ to be robust to features with mixed domains and semantics. 

While we can simply use a fixed mixing function $\phi$, as defined in Eq.~\eqref{eq:our-phi}, for Eq.~\eqref{eq:mixuploss-img} and Eq.~\eqref{eq:mixuploss-feats}, we found that it is more beneficial to devise a dynamic $\phi$ which changes its behaviour during training, in a curriculum fashion. Intuitively, minimizing the two objectives defined in Eq.\eqref{eq:mixuploss-img} and Eq.\eqref{eq:mixuploss-feats} {requires our model to correctly disentangle the various semantic components used to form the mixed samples.} While this is a complex task even for intra-domain mixes (\ie when only the semantic is mixed), mixing samples across domains makes the task even harder, requiring to isolate also domain specific factors. 
To effectively tackle this task, we choose to act on the mixing function $\phi$. {In particular, we want our $\phi$ to create mixed samples with progressively increased degree of mixing both with respect to content and domain, in a curriculum-based fashion}. 

During training we regulate both $\alpha$ (weighting the probability of cross-domain mixes) and $\beta$ (modifying the probability distribution of the mix ratio $\lambda$), changing the probability distribution of the mixing ratio $\lambda$ and of the cross-domain mix $\gamma$. In particular, given a warm-up step of $N$ epochs and being $s$ the current epoch we set $\beta=\text{min}(\frac{s}{N}\beta_{max},\beta_{max}))$, with $\beta_{max}$ as hyperparameter, while $\alpha=\text{max}(0,\text{min}(\frac{s-N}{N},1)$. As a consequence, the learning process is made of three phases, with a smooth transition among them. We start by solving the plain classification task on a single domain (\ie $s<N$,$\alpha=0$,$\beta=\frac{s}{N}\beta_{max},$). In the subsequent step ($N\leq s<2N$) samples of the same domains are mixed randomly, with possibly different semantics (\ie $\alpha=\frac{s-N}{N}$, $\beta=\beta_{max}$). In the third phase ($s\geq 2N$), we mix up samples of different domains (\ie $\alpha=1$), simulating the domain shift the predictor will face at test time. Figure \ref{fig:method}, shows a representation of how $\phi$ varies during training (top, white block).   

\myparagraph{Final objective.} {The full training procedure, is represented in Figure \ref{fig:method}. Given a training sample $(x_i,y_i,d_i)$, we randomly draw other two samples, $(x_j,y_j,d_j)$ and $(x_k,y_k,d_k)$, with $d_i=d_k,i\neq k$ and $d_j\neq d_i$, feed them to $\phi$ and obtain the first mixed input. We then feed $x_i$, $x_j$, $x_k$ and the mixed sample through $f$, to extract their respective features. At this point we use features extracted from other two randomly drawn samples (in the figure, and just for simplicity, $x_j$ and $x_k$ with same mixing ratios $\lambda$ and $\gamma$), to obtain the feature level mixed features needed to build the objective in Eq.\eqref{eq:mixuploss-feats}. Finally, the features of $x_i$ and the two mixed variants at image and feature level, are fed to the semantic projection layer $g$, which maps them to the embedding space $\mathcal{E}$. At the same time, the labels in $\mathcal{Y}^s$ are projected in $\mathcal{E}$ through $\omega$. Finally, the objectives defined in Eq.\eqref{eq:aggregation},Eq.\eqref{eq:mixuploss-img} and Eq.\eqref{eq:mixuploss-feats} functions are then computed in the semantic embedding space. } 
Our final objective is:
  \begin{equation}
 \label{eq:final-objective}
 \mathcal{L}_{\text{CuMIX}}(\mathcal{S}) = {|\mathcal{S}|^{-1}}\sum_{\mathclap{(x_i,y_i,d_i)\in\mathcal{S}}}\mathcal{L}_{\text{AGG}}(x_i,y_i)+\\\eta_{\text{I}}\mathcal{L}_{\text{M-IMG}}(x_i,y_i,d_i)+\eta_{\text{F}}\mathcal{L}_{\text{M-F}}(x_i,y_i,d_i) 
 \end{equation}
with $\eta_{\text{I}}$ and $\eta_{\text{F}}$ hyperparameters weighting the importance of the two terms. As $\ell(x,y)$ in both $\mathcal{L}_{\text{AGG}}$, $\mathcal{L}_{\text{M-IMG}}$ and $\mathcal{L}_{\text{M-F}}$, we use the standard cross-entropy loss, even if any ZSL objective can be applied. Finally, we highlight that the optimization is performed batch-wise, thus also the sampling of the triplet considers the current batch and not the full training set $\mathcal{S}$. Moreover, while in Figure \ref{fig:method} we show for simplicity that the same samples are drawn for $\mathcal{L}_{\text{M-IMG}}$ and $\mathcal{L}_{\text{M-F}}$, in practice, given a sample, the random sampling procedure of the other two members of the triplet is held-out twice, one at the image level and one at the feature level. Similarly, the sampling of the mixing ratios $\lambda$ and cross domain factor $\gamma$ of $\phi$ is held-out sample-wise and twice, one at image level and one at feature level. As in Eq.~\eqref{eq:phi-mixup}, $\lambda$ and $\gamma$ are kept fixed across mixed inputs/features and their respective targets in the label space. 

\myparagraph{Discussion.} We present similarities between our framework with DG and ZSL methods. In particular, presenting the classifier with noisy features extracted by a non-domain specialist network, has a similar goal as the episodic strategy for DG described in \cite{li2019episodic}. On the other hand, here we sidestep the need to train domain experts by directly presenting as input to our classifier features of novel domains that we obtain by interpolating the available sources samples.  
Our method is also linked to \textit{mixup} approaches developed in DA {\cite{xu2019adversarial}}. Differently from them, we use \textit{mixup} to simulate unseen domains rather then to progressively align the source to the given target data.

Our method is also related to ZSL frameworks based on feature generation \cite{xian2018feature,xian2019fvaegan}. While the quality of our synthesized samples is lower since we do not exploit attributes for conditional generation, we have a lower computational cost. {In fact, during training we simulate the test-time semantic shift without generating samples of unseen classes. Moreover, we do not require additional training phases on the generated samples or the availability of unseen class attributes to be available beforehand.}

%% file: experiments.tex
\begin{figure}[tb]
  \centering
  \includegraphics[width=0.92\columnwidth, trim = 0 0 0 0,clip]{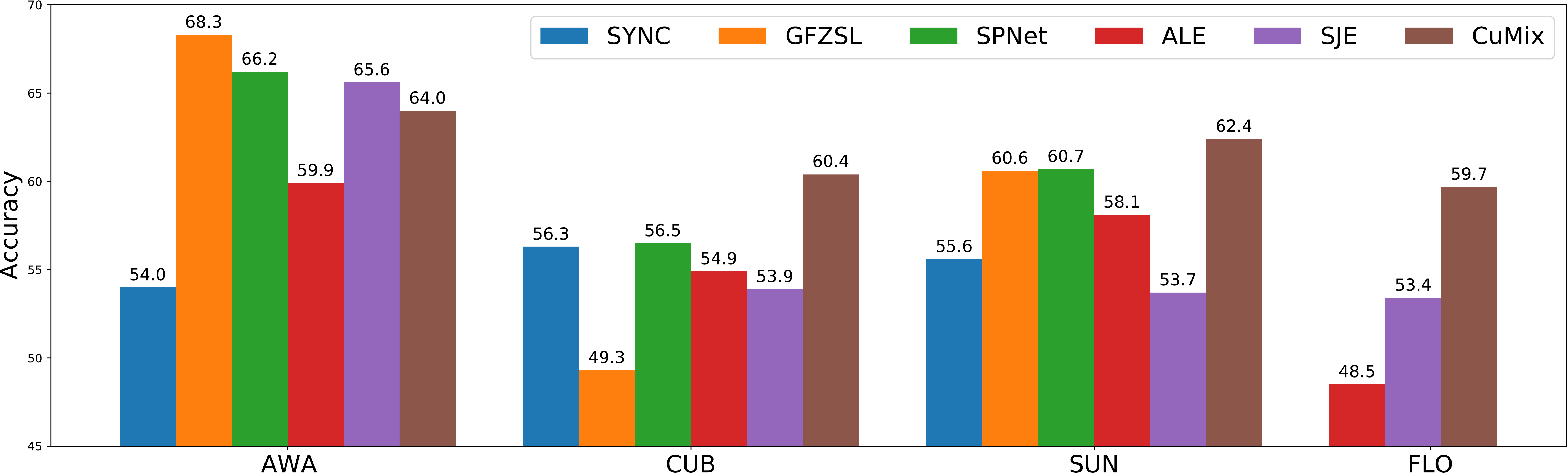} 
   \caption{ZSL results on CUB, SUN, AWA and FLO datasets with ResNet-101 features.}
  \label{fig:zsl-results}
 \end{figure}

\subsection{Datasets and implementation details}
We assess \methodName in three scenarios: ZSL, DG and the proposed ZSL+DG setting. 

\myparagraph{ZSL}. We conduct experiments on {four} standard benchmarks: Caltech-UCSD-Birds 200-2011 (CUB)~\cite{welinder2010cub}, SUN attribute (SUN)~\cite{patterson2012sun}, Animals with Attributes (AWA)~\cite{lampert2013awa} and Oxford Flowers (FLO)~\cite{nilsback2008flo}. CUB contains 11,788 images of 200 bird species, with 312 attributes, SUN 14,430 images of 717 scenes annotated with 102 attributes, and AWA 30,475 images of 50 animal categories with 85 attributes. Finally, FLO is a fine-grained dataset of flowers, containing 8,189 images of 102 categories. As semantic representation, we use the visual descriptions of \cite{reed2016learning}, following 
\cite{xian2018feature,xian2019semantic}. For each dataset, we use the train, validation and test split provided by \cite{xian2018zeroshotgood}. In all the settings we employ features extracted from the second-last layer of a ResNet-101 \cite{he2016deep} pretrained on ImageNet as image representation. For \methodName, we consider $f$ as the identity function and as $g$ a simple fully connected layer, perform our version of mixup directly at the feature-level while applying our alignment loss in the embedding space. All hyperparameters have been set following \cite{xian2018zeroshotgood}.

\myparagraph{DG}. We perform experiments on the PACS dataset \cite{li2017deeper}with 9,991 images of 7 semantic classes in 4 different visual domains, \textit{art paintings}, \textit{cartoons}, \textit{photos} and \textit{sketches}. For this experiment we use the standard train and test split defined in \cite{li2017deeper}, with the same validation protocol. We use as base architecture a ResNet-18 \cite{he2016deep} pretrained on ImageNet. For our model, we consider $f$ to be the ResNet-18 while $g$ to be the identity function. We use the same training hyperparameters and protocol of \cite{li2019episodic}. 

\myparagraph{ZSL+DG}. Since no previous work addressed the problem of ZSL+DG, there is no benchmark on this task. As a valuable benchmark, we choose DomainNet~\cite{peng2019moment}, a recently introduced dataset for multi-source domain adaptation \cite{peng2019moment} with a large variety of domains, visual concepts and possible descriptions. It contains approximately 600'000 images from 345 categories and 6 domains,  \textit{clipart}, \textit{infograph}, \textit{painting}, \textit{quickdraw}, \textit{real} and \textit{sketch}. 

To convert this dataset from a DA to a ZSL scenario, we need to define an unseen set of classes. Since our method uses a network pretrained on ImageNet \cite{russakovsky2015imagenet}, the set of unseen classes can not contain any of the classes present in ImageNet following the good practices in \cite{xian2017zero}. We build our validation + test set with 100 classes that contain at least 40 images per domain and that has no overlap with ImageNet. We reserve 45 of these classes for the unseen test set, matching the number used in \cite{thong2019open}, and the remaining 55 classes for the unseen validation set. The remaining 245 classes are used as seen classes during training. 

We set the hyperparameters of each method by training on all the images of the seen classes on a \textit{subset} of the source domains and validating on all the images of the validation set from the held-out source domain. After the hyperparameters are set, we retrain the model on the training set, i.e. 245 classes, and validation set, i.e. 55 classes, of a total number of 300 classes. Finally, we report the final results on the 45 unseen classes. As semantic representation we use word2vec embeddings \cite{mikolov2013efficient} extracted from the Google News corpus and \textit{L2}-normalized, following \cite{thong2019open}.  For all the baselines and our method, we employ as base architecture a ResNet-50 \cite{he2016deep} pretrained on ImageNet, using the same number of epochs and SGD with momentum as optimizer, with the same hyperparameters of \cite{thong2019open}.

\subsection{Results}

\myparagraph{ZSL.} 
In the ZSL scenario, we choose as baselines standard inductive methods plus more recent approaches. In particular we report the results of ALE~\cite{akata2013label}, SJE~\cite{akata2015evaluation}, SYNC~\cite{changpinyo2016synthesized}, GFZSL~\cite{verma2017simple} and SPNet~\cite{xian2019semantic}. ALE~\cite{akata2013label} and SJE~\cite{akata2015evaluation} are linear compatibility methods using a ranking loss and the structural SVM loss respectively. SYNC~\cite{changpinyo2016synthesized} learns a mapping from the feature space and the semantic embedding space by means of phantom classes and a weighted graph. GFZSL~\cite{verma2017simple} employs a generative framework where each class-conditional distribution is modeled as a multivariate Gaussian. Finally, SPNet~\cite{xian2019semantic} learns a semantic projection function from the feature space through the image embedding space by minimizing the standard cross-entropy loss. 

Our results grouped by datasets are reported in Figure \ref{fig:zsl-results}. 
Our model achieves performance either superior or comparable to the state-of-the-art in all benchmarks but AWA. We believe that in AWA learning a better alignment between visual features and attributes may not be as effective as improving the quality of the visual features. Especially, although the names of the test classes do not appear in the training set of ImageNet, for AWA being a non-fine-grained dataset, the information content of the test classes is likely represented by the ImageNet training classes. Moreover, for non-fine-grained datasets, finding labeled training data may not be as challenging as it is in fine-grained datasets. Hence, we argue that zero-shot learning is of higher practical interest in fine-grained settings. Indeed our proposed model is effective in fine-grained scenarios (\ie CUB, SUN, FLO) where it consistently outperforms the state-of-the-art approaches.

\begin{table*}[t]
		\caption{Domain Generalization accuracies on PACS with ResNet-18.} 
		\centering
		{\small
		\begin{tabular}{ p{4em}| >{\centering}p{3.7em} >{\centering}p{3.7em} >{\centering}p{3.7em} >{\centering}p{4.6em} >{\centering}p{4.3em} >{\centering}p{3.7em} >{\centering}p{4em} | >{\centering\arraybackslash}p{3.7em}}
		&AGG&DANN&MLDG&CrossGrad& MetaReg&JiGen&Epi-FCR&\methodName\\
		Target&&\cite{ganin2016domain}&\cite{li2018learning}&\cite{shankar2018generalizing}&\cite{balaji2018metareg}&\cite{carlucci2019domain}&\cite{li2019episodic}&\\
		\hline
		Photo   &  94.9 & 94.0  & 94.3  &94.0   &94.3   & \textbf{96.0} &93.9 & 95.1\\   
		Art & 76.1  &81.3   &79.5   &78.7   &79.5   &79.4   &82.1   &\textbf{82.3}\\
		Cartoon & 73.8&73.8& \textbf{77.3}&73.3&75.4&75.3&77.0&76.5\\
		Sketch &69.4&\textbf{74.3}&71.5&65.1&72.2&71.4&73.0&72.6\\
		\hline
		Average&78.5&80.8&80.7&80.7&77.8&80.4&81.5&\textbf{81.6}
		\end{tabular}
		\label{tab:dg-results}}
\end{table*}

These results show that our model based on \textit{mixup} achieves competitive performances on ZSL by simulating the semantic shift the classifier will experience at test time. To this extent, our approach is the first to show that mixup can be a powerful regularization strategy for the challenging ZSL setting.

\myparagraph{DG.} The second series of experiments consider the standard DG scenario. Here we test our model on the PACS dataset using a ResNet-18 architecture. As baselines for DG we consider the standard model trained on all source domains together (AGG), the adversarial strategies in \cite{ganin2016domain} (DANN) and \cite{shankar2018generalizing}, the meta learning-based strategy MLDG \cite{li2018learning} and MetaReg \cite{balaji2018metareg}. Moreover we consider the episodic strategy presented in \cite{li2019episodic} (Epi-FCR). 

As shown in Table \ref{tab:dg-results}, our model achieves competitive results comparable to the state-of-the-art episodic strategy Epi-FCR \cite{li2019episodic}. Remarkable is the gain obtained with respect to the adversarial augmentation strategy CrossGrad \cite{shankar2018generalizing}. Indeed, synthesizing novel domains for domain generalization is an ill-posed problem, since the concept of unseen domain is hard to capture. However, with \methodName we are able to simulate inputs/features of novel domains by simply interpolating the information available in the samples of our sources. Despite containing information available in the original sources, our approach allows to produce a model more robust to domain shift. 

Another interesting comparison is against the self-supervised approach JiGen \cite{carlucci2019domain}. Similarly to \cite{carlucci2019domain} we employ an additional task to achieve higher generalization abilities to unseen domains. While in \cite{carlucci2019domain} the JigSaw puzzles \cite{noroozi2016unsupervised} are used as a secondary self-supervised task, here we employ the mixed samples/features in the same manner. 
The improvement in the performances of our method highlights that recognizing the semantic of mixed samples acts as a more powerful secondary task to improve robustness to unseen domains.

Finally, it is worth noting that \methodName performs a form of episodic training, similar to Epi-FCR \cite{li2019episodic}. However, while Epi-FCR considers multiple domain-specific architectures (to simulate the domain experts needed to build the episodes), we require a single domain agnostic architecture. We build  
our episodes by making the \textit{mixup} among images/features of different domains increasingly more drastic. Despite not requiring any domain experts, \methodName achieves comparable performances to Epi-FCR, showing the efficacy of our strategy to simulate unseen domain shifts.

\myparagraph{Ablation study.} In this section, we ablate the various components of our method. We performed the ablation on the PACS benchmark for DG, since this allows us to show how different choices act on the generalization to unseen domains. In particular, we ablate the following implementation choices: 1) mixing samples at the image level, feature level or both 2) impact of our curriculum-based strategy for mixing features and samples. 

As shown in Table \ref{tab:ablation-study-pacs}, mixing samples at feature level produces a clear gain on the results with respect to the baseline, while mixing samples only at image level can even harm the performance. This happens particularly in the \textit{sketch} domain, where mixing samples at feature level produces a gain of ~2\% while at image level we observe a drop of ~10\% with respect to the baseline. This could be explained by mixing samples at image level producing inputs that are too noisy for the network and not representative of the actual shift experienced at test time. Mixing samples at feature level instead, after multiple layers of abstractions, allows to better synthesize the information contained in the different samples, leading to more reliable features for the classifier. Using both of them allows to obtain higher results in almost all domains. 

Finally, we analyze the impact of the curriculum-based strategy for mixing samples and features.  
As the table shows, adding the curriculum strategy allows to boost the performances for the most difficult cases (i.e. sketches) producing a further accuracy boost. Moreover, applying this strategy allows to stabilize the training procedure, as demonstrated experimentally. 

\begin{table*}[t]
			\caption{Ablation on PACS dataset with ResNet-18 as backbone.}
		\centering
		{\small
		\begin{tabular}{ c c c c | c  c  c  c | c}
		$\mathcal{L}_{\text{AGG}}$ &$\mathcal{L}_{\text{M-IMG}}$&$\mathcal{L}_{\text{M-F}}$&Curriculum&Art&Cartoon&Photo&Sketch&Avg.\\
		\hline
		\ding{51} & & & & 76.1  & 73.8  & 94.9 & 69.4 & 78.5  \\\hline
		 \ding{51}& \ding{51} & &   &78.4 &72.7   &94.7  &59.5  & 76.3  \\
	\ding{51}  &  & \ding{51} & &81.8&\textbf{76.5} &{94.9}  &71.2  &81.1  \\ 
	 \ding{51}& \ding{51} & \ding{51}   & & \textbf{82.7}& 75.4& \textbf{95.4} & 71.5 & 81.2\\ \hline
    	\ding{51}	 &\ding{51} &\ding{51}  & \ding{51}&82.3&\textbf{76.5}& 95.1 & \textbf{72.6} &\textbf{81.6} \\ 
		\end{tabular}
		}
		\label{tab:ablation-study-pacs}
\end{table*}

\myparagraph{ZSL+DG.} On the proposed ZSL+DG setting we use the DomainNet dataset, training on five out of six domains and reporting the average per-class accuracy on the held-out one. We report the results for all possible target domains but one, \ie real photos, since our backbone has been pretrained on ImageNet, thus the photo domain is not an unseen one. Since no previous methods addressed the ZSL+DG problem, in this work we consider simple baselines derived from the literature of both ZSL and DG. The first baseline is a standard ZSL model without any DG algorithm (i.e. the  standard AGG): as ZSL method we consider SPNet \cite{xian2019semantic}. The second baseline is a DG approach coupled with a ZSL algorithm. To this extent we select the state-of-the-art Epi-FCR as the DG approach, coupling it with SPNet. As a reference, we also evaluate the performance of standard \textit{mixup} coupled with SPNet.

As shown in Table \ref{tab:domainnet}, our method achieves competitive performances in ZSL+DG setting when compared to a state-of-the-art approach for DG (Epi-FCR) coupled with a state-of-the-art one for ZSL (SPNet), outperforming this baseline in almost all settings but \textit{sketch} and, in average by almost $1\%$. Particularly interesting are the results on the \textit{infograph} and \textit{quickdraw} domains. These two domains are the ones where the shift is more evident as highlighted by the lower results of the baseline. In these scenarios, our model consistently outperforms the competitors, with a remarkable gain of more than 1.5\% in average accuracy per class with respect to the ZSL only baseline. We want to highlight also that DomainNet is a challenging dataset, where almost all standard DA approaches are ineffective or can even lead to negative transfer \cite{peng2019moment}. Our method however is able to overcome the unseen domain shift at test time, improving the performance of the baselines in all scenarios. Our model consistently outperforms SPNet coupled with the standard \textit{mixup} strategy in every scenario. This demonstrates the efficacy of the choices in \methodName for revisiting \textit{mixup} in order to recognize unseen categories in unseen domains.

\begin{table*}[t]
			\caption{ZSL+DG scenario on the DomainNet dataset with ResNet-50 as backbone.} 
		\centering
		
		{\small
		\begin{tabular}{ l |  c  c  c  c  c | c   }
		Method&Clipart&Infograph&Painting&Quickdraw&Sketch&Avg.\\
		\hline
                             SPNet    &{26.0}  &16.9  & 23.8 & 8.2  & 21.8  &19.4 \\
		                           \textit{mixup}+SPNet    &27.2  &16.9  & 24.7 & 8.5  & 21.3  & 19.7\\
		
        \multirow{1}{*}{Epi-FCR+SPNet}
                               &{26.4} & 16.7  & 24.6  & 9.2 & \textbf{23.2}    &20.0\\
        \hline
        \methodName    
                                     &\textbf{27.6} &\textbf{17.8}  & \textbf{25.5}  & \textbf{9.9}  &{22.6}  & \textbf{20.7}  \\
		\end{tabular}}
		\label{tab:domainnet}
\end{table*}

%% file: conclusions.tex
In this work, we propose the novel ZSL+DG scenario. In this setting, during training,
{we are given a set of images of multiple domains and semantic categories and our goal is to build a model able to recognize unseen concepts, as in ZSL, in unseen domains, as in DG.} To solve this problem we design CuMix, the first algorithm which can be holistically and effectively applied to DG, ZSL and ZSL+DG. \methodName is based on simulating inputs and features of new domains and categories during training by mixing the available source domains and classes, both at image and feature level. 
Experiments on public benchmarks show the effectiveness of CuMix, achieving state-of-the-art performances in almost all settings in all tasks. Future works will investigate the use of alternative data-augmentation schemes in the ZSL+DG setting.

\myparagraph{\newline Acknowledgments}
We thank the ELLIS Ph.D. student program and the ERC grants 637076 - RoboExNovo (B.C.) and 853489 - DEXIM (Z.A.). This work has been partially funded by the DFG under Germany’s Excellence Strategy – EXC number 2064/1 – Project number 390727645.

%% file: suppmat/hyperparams.tex
In this section, we will detail the hyperparameter choices and validation protocols that, for lack of space, we did not include in the main paper.

\myparagraph{ZSL}. 
For each dataset, we use the train, validation and test split provided by \cite{xian2018zeroshotgood}. In all the settings we employ features extracted from the second-last layer of a ResNet-101 \cite{he2016deep} pretrained on ImageNet as image representation, \textit{without} end-to-end training. For \methodName, we consider $f$ as the identity function and as $g$ a simple fully connected layer, performing the mixing directly at the feature-level while applying our alignment loss in the embedding space (\ie $\mathcal{L}_{\text{M-IMG}}$ and $\mathcal{L}_{\text{M-F}}$ coincide in this case and are applied only once.) All hyperparameters have been set dataset-wise following \cite{xian2018zeroshotgood}, using the available validation sets. For all the experiments, we use SGD as optimizer with an initial learning rate equal to 0.1, momentum equal to 0.9, a weight-decay set to 0.001 for all settings but AWA, where is set 0. The learning-rate is downscaled by a factor of ten after 2/3 of the total number of epochs and $N=30$. In particular, for CUB and FLO we train our model for $90$ epochs, setting $\beta_{\text{max}}=0.8$ and $\eta_{\text{I}}=\eta_{\text{F}}=10.0$ for CUB, and $\beta_{\text{max}}=0.4$ and $\eta_{\text{I}}=\eta_{\text{F}}=4.0$ for FLO. For AWA, we train our network for $30$ epochs, with $\beta_{\text{max}}=0.2$ and $\eta_{\text{I}}=\eta_{\text{F}}=1.0$. For SUN, we train our network for $60$ epochs, with $\beta_{\text{max}}=0.8$ and $\eta_{\text{I}}=\eta_{\text{F}}=10$. In all settings, the batch-size is set to 128.

\myparagraph{DG.}  We use as base architecture a ResNet-18 \cite{he2016deep} pretrained on ImageNet. For our model, we consider $f$ to be the ResNet-18, $g$ to be the identity function and $\omega$ will be a learned, fully-connected classifier. We use the same training hyperparameters and protocol of \cite{li2019episodic}, setting $\beta_{\text{max}}=0.6$, $\eta_{\text{I}}=0.1$, $\eta_{\text{F}}=3$ and $N=10$.

\myparagraph{ZSL+DG.} For all the baselines and our method we employ as base architecture a ResNet-50 \cite{he2016deep} pretrained on ImageNet, using SGD with momentum as optimizer, with a learning rate of $0.001$ for the ZSL classifier and $0.0001$ for the ResNet-50 backbone, a weight decay of $5\cdot10^{-5}$ and momentum $0.9$. We train the models for 8 epochs (each epoch counted on the smallest source dataset), with a batch-size containing 24 sample per domain. We decrease the learning rates by a factor of $10$ after 6 epochs. For our model, we consider the backbone as $f$ and a simple fully-connected layer as $g$. We set $N=2$, $\eta_{\text{I}}=10^{-3}$ for all the experiments, while $\beta_{\text{max}}$ in $\{1,2\}$ and $\eta_{\text{F}}$ in $\{0.5,1,2\}$ depending on the scenario.

%% file: suppmat/zsl-dg.tex
 In Table 3 of the main paper, we showed the performance of our method in the new ZSL+DG scenario on the DomainNet dataset \cite{peng2019moment}, comparing it with three baselines: SPNet \cite{xian2019semantic}, simple \textit{mixup} \cite{zhang2017mixup} coupled with SPNet and SPNet coupled with EpiFCR \cite{li2019episodic}, an episodic-based method for DG. We reported the results of these baselines to show 1) the performance of a state-of-the-art ZSL method (SPNet), 2) the impact of \textit{mixup} alone (\textit{mixup}+SPNet) and 3) the results obtained by coupling state-of-the-art models for DG and for ZSL together (EpiFCR+SPNet). We chose SPNet and EpiFCR as state-of-the-art references for ZSL and DG respectively due to their high performances on their respective scenarios, plus because they are very recent approaches. 

In this section, we motivate our choices by showing that other baselines of ZSL and DG achieve lower performances in this new scenario. In particular we show the performances of two standard ZSL methods, ALE \cite{akata2013label} and DEVISE \cite{frome2013devise} and a standard DG/DA method, DANN \cite{ganin2016domain}. We choose DANN since it is a strong baseline for DG on residual architectures, as shown in \cite{li2019episodic}. As in the main paper, we show the performances of the ZSL methods alone, ZSL methods coupled with DANN, and with EpiFCR. For all methods, we keep the same training hyperparameters, tuning only the method-specific ones. The results are reported in Table \ref{tab:domainnet-additional}. As the table shows, \methodName achieves superior performances even compared to these new baselines. Moreover, these baselines achieve lower results than the EpiFCR method coupled with SPNet, as expected. This motivates our choices of the main paper. It is also worth highlighting how coupling ZSL methods with DANN for DG achieves lower performances than the ZSL methods alone in this scenario. This is in line with the results reported in \cite{peng2019moment}, where standard domain alignment-based methods are shown to be not effective in the DomainNet dataset, leading also to negative transfer in some cases \cite{peng2019moment}.

\begin{table*}[t]
			\caption{ZSL+DG scenario on the DomainNet dataset with ResNet-50 as backbone.} 
		\centering
		\begin{tabular}{ >{\centering}p{6em} | >{\centering}p{6em} | >{\centering}p{4em} >{\centering}p{4em} >{\centering}p{4em} >{\centering}p{4.em} >{\centering}p{4.em}| >{\centering\arraybackslash}p{4em} }
		\multicolumn{2}{c |}{Method} & \multicolumn{5}{c |}{Target Domain} &\\
		DG & ZSL &clipart&infograph&painting&quickdraw&sketch&avg.\\
		\hline
                      \multirow{3}{*}{-}&       DEVISE~\cite{frome2013devise}    & 20.1 &11.7  &17.6  &6.1   &16.7  &14.4 \\
                     &        ALE~\cite{akata2013label}    & 22.7&12.7  &20.2  &6.8  &18.5   &16.2 \\
                      &       SPNet~\cite{xian2019semantic}     &{26.0}  &16.9  & 23.8 & 8.2  & 21.8  &19.4 \\
                             
                             		\hline
		
                             \multirow{3}{*}{DANN~\cite{ganin2016domain}}&DEVISE~\cite{frome2013devise}    & 20.5&10.4&16.4&7.1&15.1&13.9 \\
                             
                             &ALE~\cite{akata2013label}     & 21.2&12.5&19.7&7.4&17.9&15.7 \\
         &SPNet~\cite{xian2019semantic}&25.9&15.8&24.1&8.4&21.3&19.1\\
		\hline
		
                             \multirow{3}{*}{EpiFCR~\cite{li2019episodic}}&DEVISE~\cite{frome2013devise}    & 21.6& 13.9 &19.3  &7.3  &17.2   &15.9 \\
                             
                             &ALE~\cite{akata2013label}     & 23.2&  14.1&21.4  &7.8  &20.9   &17.5 \\
       &SPNet~\cite{xian2019semantic} 
                               &{26.4} & 16.7  & 24.6  & 9.2 & \textbf{23.2}    &20.0\\
        \hline
        \multicolumn{2}{c |}{\methodName}    
                                     &\textbf{27.6} &\textbf{17.8}  & \textbf{25.5}  & \textbf{9.9}  &{22.6}  & \textbf{20.7}  \\
		\end{tabular}
		\label{tab:domainnet-additional}
\end{table*}

Finally, we want to underline that coupling EpiFCR with any of the ZSL baselines, is not a straightforward approach, but requires to actually adapt this method, re-structuring the losses. In particular, we substitute the classifier originally designed for EpiFCR with the classifier specific of the ZSL method we apply on top of the backbone. Moreover, we additionally replace the classification loss with the loss devised for the particular ZSL method. For instance, for EpiFCR+SPNet, we use as classifier the semantic projection network, using the cross-entropy loss in \cite{xian2019semantic} as classification loss. Similarly, for EpiFCR+DEVISE and EpiFCR+ALE, we use as classifier a bi-linear compatibility function \cite{xian2018zeroshotgood} coupled with a pairwise ranking objective \cite{frome2013devise} and with a weighted pairwise ranking objective \cite{akata2013label} respectively.

%% file: suppmat/zsl-ablation.tex
In this section, we report the ZSL results in tabular form. The results are shown in Table \ref{tab:zsl-results-additional}. 
With respect to Figure 3 of the main paper, in the table, we also report the results of a baseline which uses just the cross-entropy loss term (similarly to \cite{xian2019semantic}), without the mixing term employed in our \methodName method. As the table shows, our baseline is weak, performing below most of the ZSL methods in all scenarios but FLO. However, adding our mixing strategy allows to boost the performances in all scenarios, achieving state-of-the-art performances in most of them. We also want to highlight that in Table \ref{tab:zsl-results-additional}, as in the main paper, we do not report the results of methods based on generating features of unseen classes for ZSL \cite{xian2018feature,xian2019fvaegan}. This choice is linked to the fact that these methods can be used as data augmentation strategies to improve the performances of any ZSL method, as shown in \cite{xian2018feature}. While using them can improve the results of all the baselines as well as \methodName, this falls out of the scope of this work.

\begin{table*}[t]
			\caption{ZSL results.}
		\centering
		\begin{tabular}{ l |  c  c  c  c}
		Method&CUB&SUN&AWA1&FLO\\
		\hline
		ALE  \cite{akata2013label}   & 54.9  & 58.1  & 59.9& 48.5   \\
		SJE   \cite{akata2015evaluation}  &  53.9 & 53.7  &   65.6& 53.4 \\
		SYNC   \cite{changpinyo2016synthesized}  &  56.3 & 55.6   & 54.0& -  \\
		GFZSL \cite{verma2017simple}    &  49.3 & 60.6  & \textbf{68.3}&  -  \\
		SPNet \cite{xian2019semantic}    & 56.5  & 60.7  & 66.2& -  \\
		\hline
		Baseline  & 52.4  & 58.2 & 62.5 & 58.4 \\
		\methodName      & \textbf{{60.4}} & \textbf{62.4} & 64.0  &\textbf{ 59.7 }\\
		\end{tabular}
		\label{tab:zsl-results-additional}
\end{table*}